\documentclass[journal]{IEEEtran}
\usepackage[style=ieee, backend=biber, maxnames=2, minnames=1]{biblatex}
\ifCLASSINFOpdf
\else
\fi
\usepackage{amsmath,amssymb,amsfonts}
\usepackage{lettrine}
\usepackage{amsfonts,amssymb}
\usepackage{amsfonts}
\usepackage{url}
\usepackage{balance}
\usepackage{soul}
\usepackage{multirow}
\usepackage{colortbl} 
\usepackage{xcolor} 
\usepackage[ruled,linesnumbered]{algorithm2e}
\usepackage{algpseudocode}
\usepackage{amsmath}
\usepackage{graphics}
\usepackage{epsfig}
\usepackage{color}
\usepackage[colorlinks,
            linkcolor=blue,
            anchorcolor=blue,
            citecolor=blue,
            urlcolor=blue]{hyperref}
\usepackage{float}
\usepackage{verbatim}
\usepackage{orcidlink}
\usepackage{multirow}
\usepackage{color}
\usepackage{colortbl,hhline}
\usepackage{xcolor}
\usepackage{colortbl}
\usepackage{nomencl}
\makenomenclature
\usepackage{soul}
\usepackage{array}
\usepackage{diagbox}
\usepackage{arydshln}
\usepackage[T1]{fontenc}
\usepackage{booktabs}
\usepackage{makecell} 
\usepackage{booktabs} 
\usepackage{multirow}
\usepackage[normalem]{ulem}

\useunder{\uline}{\ul}{}
\title{Advancing Embodied Intelligence in Robotic-Assisted Endovascular Procedures: \\ A Systematic Review of AI Solutions}\vspace{-1.5cm}

\begin{document}

\author{Tianliang Yao$^{1,2}$, Bo Lu$^{3}$, Markus Kowarschik$^{4}$, Yixuan Yuan$^{2}$, Hubin Zhao$^{5}$, Sebastien Ourselin$^{6}$, \\Kaspar Althoefer$^{7}$, Junbo Ge$^{8}$, and Peng Qi$^{1,9,*}$\vspace{-1.0cm}
\thanks{The authors would like to thank Mr. Mish Toszeghi from Queen Mary University of London for his meticulous proofreading of this manuscript. The authors also appreciate Mr. Haoyu Wang from Shanghai University of Engineering Science for his constructive discussions and valuable suggestions. This work is supported by the National Key Research and Development Program of China under Grant No. 2023YFB4705200, the National Natural Science Foundation of China under Grant No. 62273257, and the Open Project Fund of State Key Laboratory of Cardiovascular Diseases No. 2024SKL-TJ002. \emph{*Corresponding author: Peng Qi (pqi@tongji.edu.cn)}}
\thanks{$^{1}$Department of Control Science and Engineering, College of Electronics and Information Engineering, and Shanghai Institute of Intelligent Science and Technology, Tongji University, Shanghai 200092, China;}
\thanks{$^{2}$Department of Electronic Engineering, Faculty of Engineering, The Chinese University of Hong Kong, Hong Kong SAR 999077, China;}
\thanks{$^{3}$Robotics and Microsystems Center, School of Mechanical and Electrical Engineering, Soochow University, Suzhou 215131, China;}
\thanks{$^{4}$Siemens Healthineers Advanced Therapies (AT), Forchheim, Bavaria 91301, Germany;}
\thanks{$^{5}$HUB of Intelligent Neuro-Engineering (HUBIN), CREATe, Division of Surgery \& Interventional Science, University College London, London, HA7 4LP, UK;}
\thanks{$^{6}$School of Biomedical Engineering \& Imaging Sciences, King's College London, London SE1 7EH, UK;}
\thanks{$^{7}$Centre for Advanced Robotics @ Queen Mary, School of Engineering and Materials Science, Queen Mary University of London, London E1 4NS, UK;}
\thanks{$^{8}$Department of Cardiology, Zhongshan Hospital, Fudan University, Shanghai Institute of Cardiovascular Diseases, Shanghai 200032, China;}
\thanks{$^{9}$State Key Laboratory of Cardiovascular Diseases and Medical Innovation Center, Shanghai East Hospital, School of Medicine, Tongji University, Shanghai 200092, China.}}


\markboth{Preprint Submitted to IEEE}%
{Yao \MakeLowercase{\textit{et al.}}: Advancing}

\maketitle

\begin{abstract}
Endovascular procedures have revolutionized vascular disease treatment, yet their manual execution is challenged by the demands for high precision, operator fatigue, and radiation exposure. Robotic systems have emerged as transformative solutions to mitigate these inherent limitations. A pivotal moment has arrived, where a confluence of pressing clinical needs and breakthroughs in AI creates an opportunity for a paradigm shift toward Embodied Intelligence (EI), enabling robots to navigate complex vascular networks and adapt to dynamic physiological conditions. Data-driven approaches, leveraging advanced computer vision, medical image analysis, and machine learning, drive this evolution by enabling real-time vessel segmentation, device tracking, and anatomical landmark detection. Reinforcement learning and imitation learning further enhance navigation strategies and replicate expert techniques. 
This review systematically analyzes the integration of EI into endovascular robotics, identifying profound systemic challenges such as the heterogeneity in validation standards and the gap between human mimicry and machine-native capabilities. Based on this analysis, a conceptual roadmap is proposed that reframes the ultimate objective away from systems that supplant clinical decision-making. This vision of augmented intelligence, where the clinician's role evolves into that of a high-level supervisor, provides a principled foundation for the future of the field.
\end{abstract}

\vspace{-0.19cm}


\vspace{-0.3cm}


\section{Introduction}
\begin{figure*}[!htbp]
    \centering
    \includegraphics[width=0.94\textwidth, keepaspectratio]{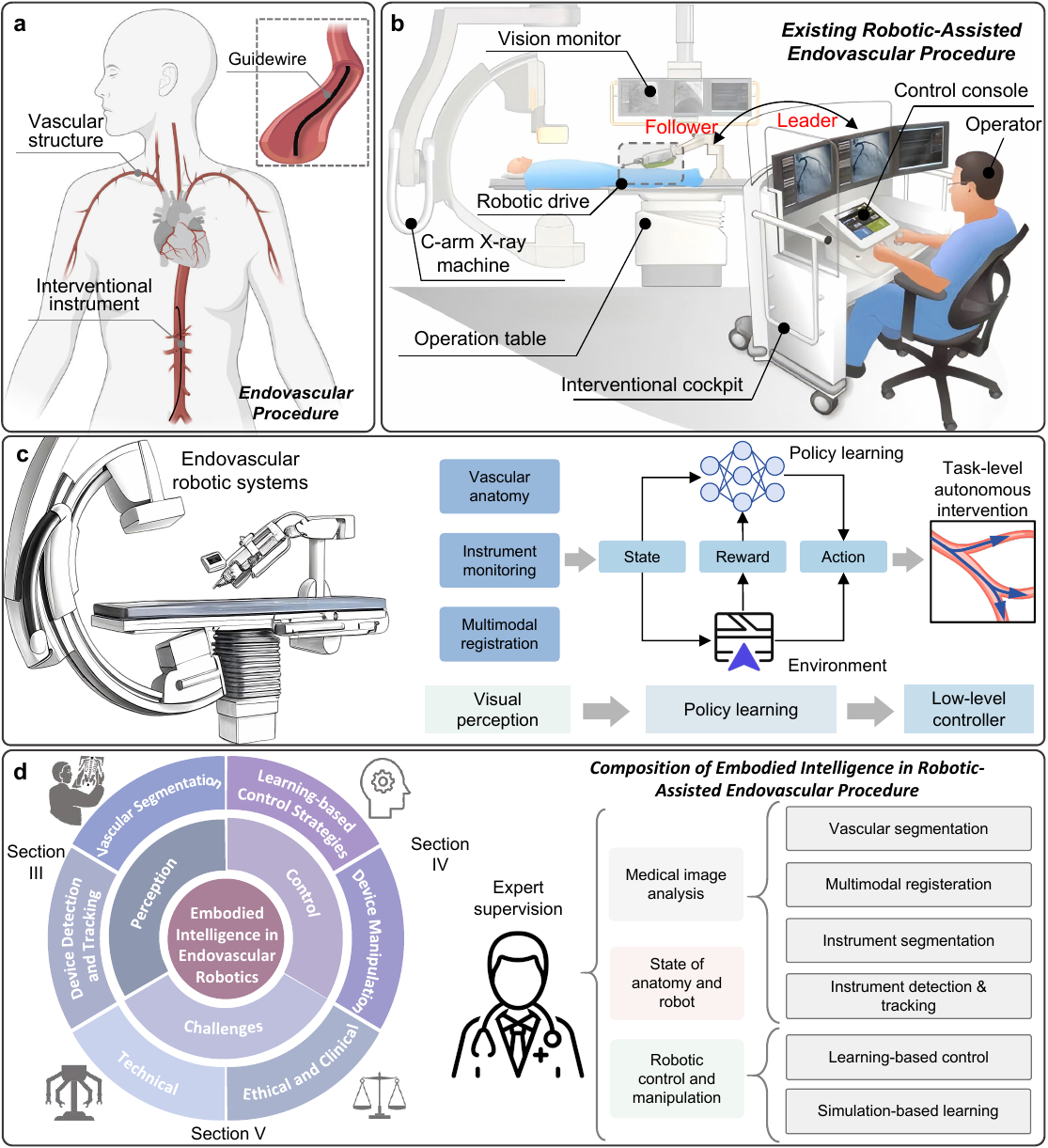}
    \vspace{-0.3cm}
    \caption{Overview of endovascular intervention, robotic assistance, and the proposed EI framework. (a) Endovascular procedure: flexible instruments (guidewires and catheters) are introduced percutaneously via radial or femoral access and navigated within the vascular lumen under X‑ray fluoroscopy to reach targets such as coronary stenoses or aneurysms for diagnosis and therapy. (b) Robotic-assisted workflow: a leader–follower architecture is depicted in which the operator at a cockpit (leader console) issues push–pull, rotate, and torque commands that are executed by a bedside robotic drive (follower) manipulating the instruments. A C‑arm X‑ray system provides real-time imaging, and a vision monitor displays imaging and system status. The operating table supports the patient; optional sensors (e.g., encoders, force/torque, haptics) enable safety monitoring and feedback. (c) Conceptual framework for EI in robotic-assisted endovascular procedures: perception modules (vascular anatomy segmentation, instrument monitoring, and multimodal registration with preoperative imaging) form the system state; a decision layer uses learning-based policy optimization (e.g., reinforcement learning) with task models and safety constraints to select high-level actions; a low-level controller executes device motions subject to hardware limits. Training and evaluation occur in both simulation and the real operating room; a patient-specific digital twin can model anatomy, device-vessel interaction, imaging geometry, policy learning, and online adaptation. Arrows indicate information and control flow. (d) Composition of EI: core capabilities include vascular segmentation and topology extraction, device detection, tracking, and tip localization, multimodal 2D/3D registration for spatial context, and learning-based control for navigation tasks such as branch cannulation, lesion crossing, and device delivery. Key technical challenges include real-time integration, generalization across vendors and anatomies, robustness to motion and low contrast, and uncertainty estimation; ethical and clinical considerations include patient safety, radiation management, data privacy, and regulatory compliance, with expert-in-the-loop supervision for safe autonomy. The schematic was created with BioRender (\url{https://biorender.com}).\vspace{-0.5cm}}
    \label{fig:Framework}
\end{figure*}

\vspace{-0.1cm}
Endovascular procedures have been subject to a paradigm shift in relation to the management of vascular disease with the advent of minimally invasive therapeutic strategies that significantly shorten patient recovery times and enhance clinical outcomes \cite{gaudino2023current,topol2015textbook}. Central to these procedures is the intricate manipulation of catheters and guidewires under real-time imaging guidance, a task that necessitates exceptional technical skill and dexterity, as illustrated in Fig. \ref{fig:Framework}(a), which outlines the workflow of an endovascular procedure \cite{pore2023autonomous, konda2025robotically}. Despite unparalleled progress, conventional manual approaches are constrained by factors such as operator fatigue, radiation exposure, and the inherent shortcomings of human precision \cite{li2024robotic, robertshaw2023artificial}. The emergence of robotic systems has begun to address these challenges, providing enhanced stability, accuracy, and reproducibility, thereby augmenting the efficacy of these procedures \cite{wu2024review}. Nevertheless, current robotic systems still use leader-follower technologies, as depicted in Figure \ref{fig:Framework}(b), in which the operator manipulates the control console from within the operational cockpit. In this configuration, the robotic system translates the operator's inputs into precise movements of the catheter and guidewire, facilitating remote control and minimizing the need for direct physical interaction within the patient. This design enables remote operation and improved ergonomics, yet constrains the robotic system to passive execution without autonomous perception or decision-making, as reflected by the unidirectional, operator-to-robot command pathway in Fig. \ref{fig:Framework}(b).

To advance beyond such teleoperated architectures, the concept of EI has emerged as a pivotal theoretical and technological driver \cite{gupta2021embodied, liu2024evolution}, particularly within the medical domain \cite{xu2024transforming, xu2024enhancing}. EI diverges from traditional robotic design philosophies by emphasizing the interplay between a robot's physical embodiment and its intelligent behaviors \cite{deitke2022retrospectives, billard2025roadmap}. This is particularly relevant for endovascular robotics, in which systems must navigate complex, deformable vascular networks and be able to adapt, in real-time, to a continuously shifting physiological environment \cite{chu2023advances}. In the context of endovascular robotics, EI refers to physically instantiated agents that operate within the human vasculature, processing multimodal sensory inputs, interpreting anatomical context, and generating adaptive control outputs in real time, within the perception–decision–control architecture schematized in Fig. \ref{fig:Framework}(c) and the capability decomposition in Fig. \ref{fig:Framework}(d). By incorporating EI principles, future robotic platforms can transcend pre-defined, scripted responses, and begin moving towards adaptive, contextually aware capabilities that more closely mimic the nuanced skills of experienced physicians \cite{yip2023artificial,zhang2024ai}. As depicted in Fig. \ref{fig:Framework}(c), EI can also serve as a co-pilot, assisting physicians by providing real-time decision support and optimizing procedural strategies, closing the loop between perception and action via the arrowed information and control flows, with expert-in-the-loop oversight highlighted in Fig. \ref{fig:Framework}(c,d).

The integration of data-driven techniques provides the computational foundation for realizing EI in robotic-assisted endovascular procedures. Advanced computational methodologies, from computer vision and learning-enhanced medical image analysis to reinforcement and imitation learning, hold the potential to distill, harness and replicate the expertise of experienced clinicians. These methodologies facilitate precise anatomical interpretation, improved device tracking, and adaptive procedural planning, effectively merging the cognitive and motor elements of interventional practice into a unified, intelligent robotic architecture.

Over time, endovascular robotics has evolved from rudimentary platforms offering remote manipulation and limited motion control to sophisticated systems equipped with multi-modal sensing and intelligent decision support. With the advent of Artificial Intelligence (AI) and Machine Learning (ML), next-generation systems now strive for greater autonomy, incorporating automated path planning and real-time obstacle avoidance into their repertoires. Empirical evidence from preclinical studies underscores the promise of these developments, heralding a future in which endovascular procedures are safer, more efficient, and decreasingly operator-dependent \cite{su2022state}.

\begin{figure}[!htbp]
\centering
\includegraphics[width=0.4\textwidth]
{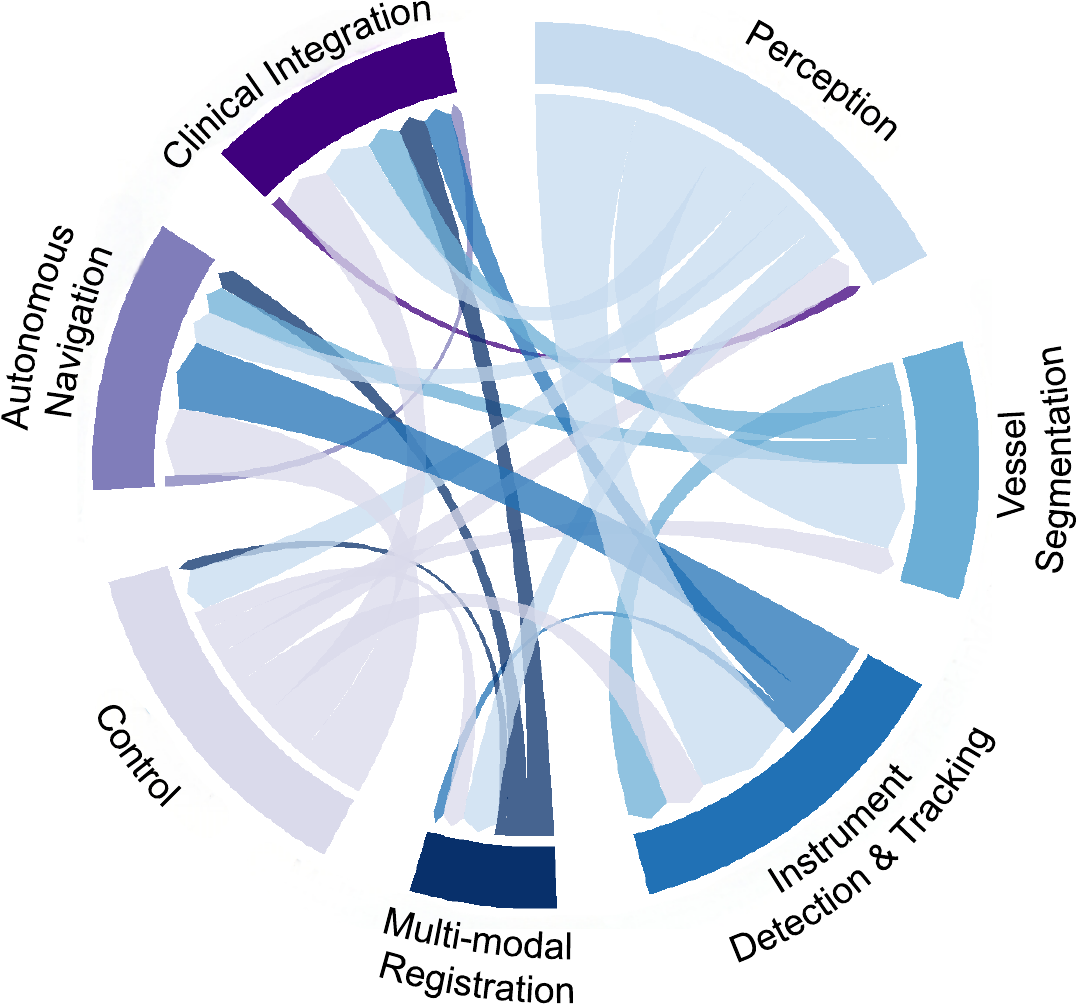}
\vspace{-0.2cm}
\caption{Chord diagram depicting the interplay between perception and control modules in robotic-assisted endovascular procedures. Perception modules (blue) include vessel segmentation, instrument detection and tracking, and multi-modal registration, linked to control modules (purple) via clinical integration. Control encompasses autonomous navigation. Arrows denote directional relationships, with line thickness reflecting connection strength, illustrating their integrated roles in enhancing procedural outcomes.\vspace{-0.2cm}}
\label{fig:Chord_diagram}
\end{figure}

Recent breakthroughs in deep learning and computer vision are accelerating progress toward this integrated perception-control pipeline. State-of-the-art algorithms now enable real-time vascular segmentation~\cite{huang2024spironet}~and device localization~\cite{du2024guidewire}, providing the rich sensory input necessary for an EI framework. Concurrently, the application of reinforcement learning is refining navigation and decision-making policies, allowing robotic agents to adapt to patient-specific vascular geometries. Hybrid solutions that combine data-driven autonomy with traditional control are also emerging, benefiting from both robust engineering principles and the flexibility of computational intelligence. The intricate interplay between these perception and control modules is foundational to EI. As our literature analysis reveals (Fig.~\ref{fig:Chord_diagram}), there is a strong co-occurrence of these components, highlighting their collaborative and synergistic role in advancing endovascular interventions.

Despite these advancements, significant challenges persist that impede the translation of these technologies into widespread clinical practice. These include the technical hurdles of achieving robust generalization, the systemic challenge of data scarcity, and the practical barriers of regulatory approval and seamless workflow integration. More fundamentally, a "translation gap" exists, rooted in a paradigm mismatch: the goal of robotics should not be to merely replicate human actions, but to leverage the unique capabilities of machines to solve clinical tasks in new ways. This review posits that clinical adoption will be most effectively accelerated not by demonstrating equivalence, but by identifying "transcendent scenarios" where embodied agents can demonstrably operate beyond the limits of human ability, thus creating a clear and compelling clinical pull.

Previous reviews have made important contributions by providing valuable analyses of distinct facets of this domain. For instance, some have offered systematic accounts of motion planning techniques for navigating guidewires within static vascular models~\cite{pore2023autonomous}. Others have focused on the application of specific AI methods like reinforcement learning for autonomous catheter navigation, with many studies predominantly validated in simulation environments~\cite{robertshaw2023artificial, wu2024review}, or have provided broader evaluations of the evolution and ethics of AI-enhanced surgical robots in general~\cite{liu2024evolution}. While insightful, these works often treat fluoroscopic image perception as a discrete computer vision problem or catheter control as an isolated robotics challenge. They have not yet provided an overarching conceptual framework that unifies these elements to address the holistic task of navigating a flexible instrument through a dynamic, pulsating vascular system.

In contrast, the present review aims to provide such a unifying perspective by analyzing the field through the integrated lens of EI. We systematically dissect the entire perception-to-control pipeline as a cohesive whole, which enables a physical robot to perceive the state of its guidewire within the vasculature and decide upon the next action, such as a push or a rotation. Our analysis is structured around the core components of this EI framework: intelligent perception for situational awareness within the vessel (Section~\ref{Intelligent Perception}), and learning-based control for instrument manipulation (Section~\ref{Control}). This is followed by a critical evaluation of the key challenges that impede clinical translation (Section~\ref{Challenges}), paying special attention to the systemic and ethical barriers that define the human-robot relationship, such as the "translation gap" and the paradigm mismatch between mimicking a surgeon's hand movements and leveraging machine-native capabilities. By synthesizing these elements, this review constructs a conceptual roadmap and argues for a future centered on sophisticated human-machine collaboration in the catheterization lab. We posit that the ultimate objective is not the pursuit of "full autonomy," but the creation of "transcendent scenarios" where an embodied agent can, for example, maintain stable guidewire placement in a tortuous distal vessel, a task where human manual dexterity often falters. This vision foresees an evolution of the expert interventionalist's role: from a master dependent on manual skill, to a high-level supervisor who directs the robot's overall strategy and intervenes at critical bifurcations. It is this focus on augmented intelligence, rather than replacement, that provides a principled foundation for guiding the future development of the field.

\vspace{-0.2cm}

\begin{figure}[!htbp]
\centering
\includegraphics[width=0.5\textwidth]{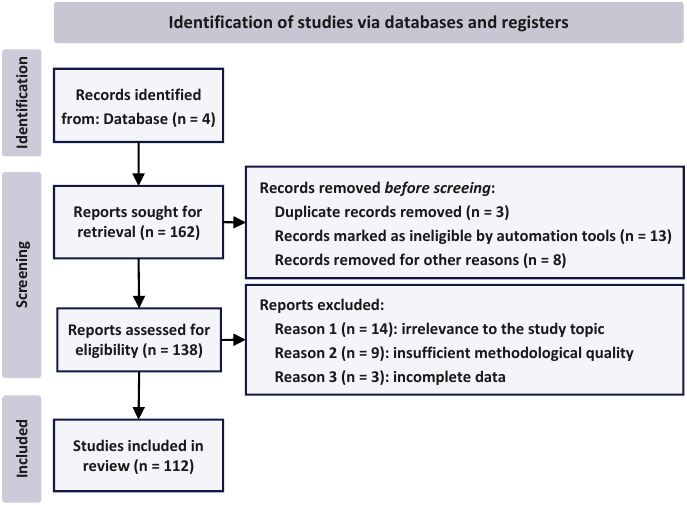}\vspace{-0.2cm}
\caption{PRISMA flowchart illustrating the study selection process. A total of 138 articles were screened for eligibility. Of these, 26 were excluded due to irrelevance to the study topic, insufficient methodological quality, or incomplete data. Ultimately, 112 studies were included in the final review.\vspace{-0.6cm}} 
\label{fig:Flowchart}
\end{figure}

\section{Search Methodology}
A systematic literature review was conducted following the Preferred Reporting Items for Systematic Reviews and Meta-Analyses (PRISMA) guidelines, as illustrated in Fig.~\ref{fig:Flowchart}. The search strategy was applied across four major scholarly databases (IEEE Xplore, PubMed, ScienceDirect, and the ACM Digital Library), utilizing combinations of phrases designed to capture three core concepts: the robotic system (e.g., "endovascular robot"), the clinical domain (e.g., "guidewire navigation"), and the intelligent methodology (e.g., "learning-based control", "embodied intelligence"). A specific emphasis was placed on articles published from 2020 to the present, a timeframe selected to capture the maturation of data-driven approaches following the articulation of evolving concepts of EI in 2019~\cite{howard2019evolving}, reflecting a shift in the field towards solving complex, real-world clinical integration challenges.

The initial search process identified 162 reports for retrieval. These reports then underwent a preliminary screening process to ensure relevance and quality before the main eligibility assessment. During this step, a total of 24 records were removed. Specifically, 3 records were removed as duplicates. An additional 13 records were excluded by automated screening tools, primarily for reasons such as being non-English language publications or not conforming to the article type (e.g., editorials or comments). A further 8 records were removed for other reasons, including being conference abstracts without a corresponding full paper or reports where the full text was irretrievable. This preliminary filtering resulted in 138 unique and potentially relevant reports being advanced to the full-text eligibility assessment stage.

In the second stage, a full-text review of these 138 articles was performed, resulting in the exclusion of an additional 26 reports. The reasons were carefully documented: 14 reports were excluded for irrelevance, such as focusing exclusively on mechanical design or new materials without an intelligent perception or control component; 9 were excluded for insufficient methodological quality, lacking the detail needed for a meaningful analysis of their algorithms; and 3 were excluded for incomplete reported data, which precluded any evaluation. This comprehensive screening process ultimately yielded 112 studies for qualitative synthesis.

A central finding that emerged from this systematic analysis is the significant heterogeneity in experimental settings, validation metrics, and anatomical targets across the included studies. Recognizing this, the synthesis was designed not as a quantitative meta-anaylsis, which would be statistically inappropriate, but as a qualitative categorization of the diverse approaches. Data extraction focused on key themes, including perception methods, control algorithms, validation environments, and reported performance metrics. This systematic approach enabled the mapping of the field's fragmented landscape, the identification of emerging trends, and the structuring of the research gaps that form the basis of this review.

\begin{table*}[!t]
\centering
\caption{Comprehensive Review of Data-Driven Methods for Endovascular Procedure Scene Perception: Clinical Applications and Challenges}
\renewcommand{\arraystretch}{1.2}
\begin{tabular}{p{3.1cm}p{2.0cm}p{6.0cm}p{2.5cm}p{2.0cm}}
\toprule
\textbf{Clinical Task} & \textbf{Authors} & \textbf{Method and Key Features} & \textbf{Performance Metrics} & \textbf{Dataset} \\
\midrule
\multirow{8}{*}{\parbox{3.1cm}{Vessel Segmentation for Diagnosis and Stenosis Assessment \\ \textit{Challenge:} Accurate delineation of coronary anatomy amidst motion artifacts and variable contrast}} & Wang \textit{et al.} \cite{wang2020coronary}, Hao \textit{et al.} \cite{hao2020sequential}, Shi \textit{et al.} \cite{shi2020uenet}, Xian \textit{et al.} \cite{xian2020main}, Li \textit{et al.} \cite{li2020cau}, Zhu \textit{et al.} \cite{zhu2021coronary}, Thuy \textit{et al.} \cite{thuy2021coronary}, Tian \textit{et al.} \cite{tian2021automatic}, Liu \textit{et al.} \cite{liu2022full}, Najarian \textit{et al.} \cite{gao2022vessel} & CNN \cite{chan2020deep} and GAN-based methods \cite{xun2022generative} with spatio-temporal and attention mechanisms; \textit{Key Features:} Temporal consistency, multi-scale feature fusion, motion artifact reduction, and morphological refinement for precise vessel delineation & IoU: 0.714--0.957, F1: 0.760--0.900, Dice: 0.764--0.939, Acc: 0.835--0.957 & Private, DCA1 \cite{cervantes2019automatic}, CHUAC \cite{carballal2018automatic} \\
\cmidrule{2-5}
& Iyer \textit{et al.} \cite{iyer2021angionet}, Song \textit{et al.} \cite{song2022coronary}, Nobre \textit{et al.} \cite{nobre2023coronary}, Shen \textit{et al.} \cite{shen2023dbcu} & Clinical validation with UNet and ConvLSTM architectures; \textit{Key Features:} Stenosis assessment, vessel width prediction, and multi-center validation for diagnostic reliability & Dice: 0.773--0.948, F1: 0.879 & Private \\
\cmidrule{2-5}
& Zhang \textit{et al.} \cite{zhang2020weakly,zhang2021ss}, Qi \textit{et al.} \cite{qi2021examinee}, Ma \textit{et al.} \cite{ma2021self}, Shen \textit{et al.} \cite{shen2023expert} & Weakly/semi-supervised methods; \textit{Key Features:} Noisy label handling, self-ensembling, and knowledge distillation to reduce annotation burden & Dice: 0.557--0.821, F1: 0.809 & Private, XCAD \cite{ma2021self}, DCA1 \cite{cervantes2019automatic} \\
\midrule
\multirow{5}{*}{\parbox{3.1cm}{Advanced Vessel Analysis for Procedural Planning \\ \textit{Challenge:} Robust feature extraction under limited annotations and complex vessel geometries}} & Hau \textit{et al.} \cite{zhang2022progressive}, Han \textit{et al.} \cite{han2022recursive}, Zhang \textit{et al.} \cite{zhang2023centerline}, Navab \textit{et al.} \cite{zhang2024self} & Centerline-aware and diffusion-based methods; \textit{Key Features:} Recursive learning, direction-aware features, and contrastive diffusion for complex vessel modeling & F1: 0.831--0.856, Dice: 0.853--0.939 & Private, MCS, DCAI \cite{cervantes2019automatic}, XCA \cite{cervantes2019automatic} \\
\cmidrule{2-5}
& Zhang \textit{et al.} \cite{zhang2023partial}, Ma \textit{et al.} \cite{ma2023towards}, Lee \textit{et al.} \cite{wu2024denver}, Kim \textit{et al.} \cite{kim2024c}, Lei \textit{et al.} \cite{lei2025enriching}, Zhang \textit{et al.} \cite{zhang2024curve} & Partial annotation and diffusion-based frameworks; \textit{Key Features:} Coarse-to-fine detection, deformable modeling, and topological connectivity for robust planning & Dice: 0.621--0.763, F1: 0.791, PCK: 0.763 & Private, XCAD \cite{ma2021self}, CADICA \cite{jimenez2024cadica}, 134XCA \cite{cervantes2019automatic}, 30XCA \cite{hao2020sequential} \\
\cmidrule{2-5}
& Wang \textit{et al.} \cite{wang2023nc}, Yang \textit{et al.} \cite{yang2024segmentation}, Guo \textit{et al.} \cite{guo2024edge} & Geometry-constrained and edge-preserving methods; \textit{Key Features:} Cross-modality learning and scribble supervision for enhanced procedural planning & Dice: 0.734--0.895, F1: 0.801 & Private, ASOCA \cite{gharleghi2022automated}, CCTA-200, XCAD \cite{ma2021self} \\
\midrule
\multirow{4}{*}{\parbox{3.1cm}{Guidewire and Catheter Tracking for Navigation \\ \textit{Challenge:} Real-time, precise localization in dynamic environments}} & Zhou \textit{et al.} \cite{zhou2020pyramid,zhou2020lightweight,zhou2020real}, Li \textit{et al.} \cite{li2021unified,li2021gaussian}, Zhu \textit{et al.} \cite{zhu2021multi}, Zhang \textit{et al.} \cite{zhang2022real} & Recurrent and attention-based tracking; \textit{Key Features:} Real-time processing, multi-scale attention, and probabilistic endpoint localization & Dice: 0.899--0.947, F1: 0.915--0.938, Acc: 0.998, AP: 0.907 & Private \\
\cmidrule{2-5}
& Zhang \textit{et al.} \cite{zhang2023jigsaw}, Ravigopal \textit{et al.} \cite{ravigopal2023real}, Wen \textit{et al.} \cite{wen2024generalizing}, Du \textit{et al.} \cite{du2024guidewire} & Transformer and sim-to-real frameworks; \textit{Key Features:} Jigsaw training, kinematic modeling, and domain adaptation for robust tracking & F1: 0.846--0.899, IoU: 0.660, Mean Error: 2.02 px & Private, Cardiac \cite{gherardini2020catheter} \\
\cmidrule{2-5}
& Gherardini \textit{et al.} \cite{gherardini2020catheter}, Kolen \textit{et al.} \cite{yang2020deep}, Zhou \textit{et al.} \cite{zhou2020frr}, Omisore \textit{et al.} \cite{omisore2021automatic}, Aghasizade \textit{et al.} \cite{aghasizade2023coordinate}, Ranne \textit{et al.} \cite{ranne2024cathflow} & Lightweight and self-supervised catheter tracking; \textit{Key Features:} Synthetic data, deep Q-learning, and optical flow for navigation & Dice: 0.650--0.950, F1: 0.950, AUC: 0.753, Mean Error: 7.36 px & Private, Cardiac \cite{gherardini2020catheter} \\
\cmidrule{2-5}
& Lalinia \textit{et al.} \cite{lalinia2024coronary}, Yao \textit{et al.} \cite{yao2023enhancing} & Edge-guided and path planning systems; \textit{Key Features:} Multi-scale edge detection and heuristic optimization for intraoperative guidance & Dice: 0.818, F1: 0.755 & Private \\
\midrule
\multirow{2}{*}{\parbox{3.1cm}{Multi-Modal Registration for Procedural Guidance \\ \textit{Challenge:} Aligning 2D / 3D data in real-time with minimal error}} & Wu \textit{et al.} \cite{wu2022car}, Yan \textit{et al.} \cite{yan20233d}, Jaganathan \textit{et al.} \cite{jaganathan2023self}, Huang \textit{et al.} \cite{huang2024real}, Liu \textit{et al.} \cite{liu2024udcr} & 3D/2D registration with reinforcement learning and CNN; \textit{Key Features:} Deformation modeling, centroid alignment, and reward-based optimization for precise alignment & Dice: 0.767--0.854, Mean Error: 1.13--2.85 mm, MPE: 2.48--2.65 & Private \\
\bottomrule
\end{tabular}
\label{tab:clinical_vessel_seg_methods}
\vspace{1em}
\footnotesize
\textit{Note:} Performance metrics include IoU (Intersection over Union), F1, Dice, Acc (Accuracy), AP (Average Precision), AUC (Area Under Curve), MPE (Mean Projective Error), and Mean Error (in pixels or mm). Metrics are reported as in the original papers.
\end{table*}

\section{Intelligent Perception Techniques in Endovascular Procedures}\label{Intelligent Perception}

Endovascular procedures, including percutaneous coronary interventions and neurovascular embolizations, demand precise, real-time interpretation of dynamic anatomical structures under imaging constraints imposed by motion artifacts, limited soft-tissue contrast, and radiation dose considerations. Intelligent perception techniques, encompassing vessel segmentation, interventional device tracking, and multi-modal registration, constitute foundational capabilities for robotic autonomy by enabling scene understanding that reduces reliance on continuous operator input. These capabilities map directly onto clinical workflow stages: preoperative planning benefits from automated vessel extraction and lesion characterization; intraoperative navigation depends on real-time instrument localization relative to vascular roadmaps; device deployment requires fusion of preoperative three-dimensional anatomical models with live two-dimensional fluoroscopy; and postoperative assessment leverages automated detection of residual stenosis or adverse events. Despite substantial algorithmic progress, clinical translation remains constrained by fundamental challenges in generalization, reproducibility, and robustness.

\begin{figure*}[!t]
\centering
\includegraphics[width=0.86\textwidth]{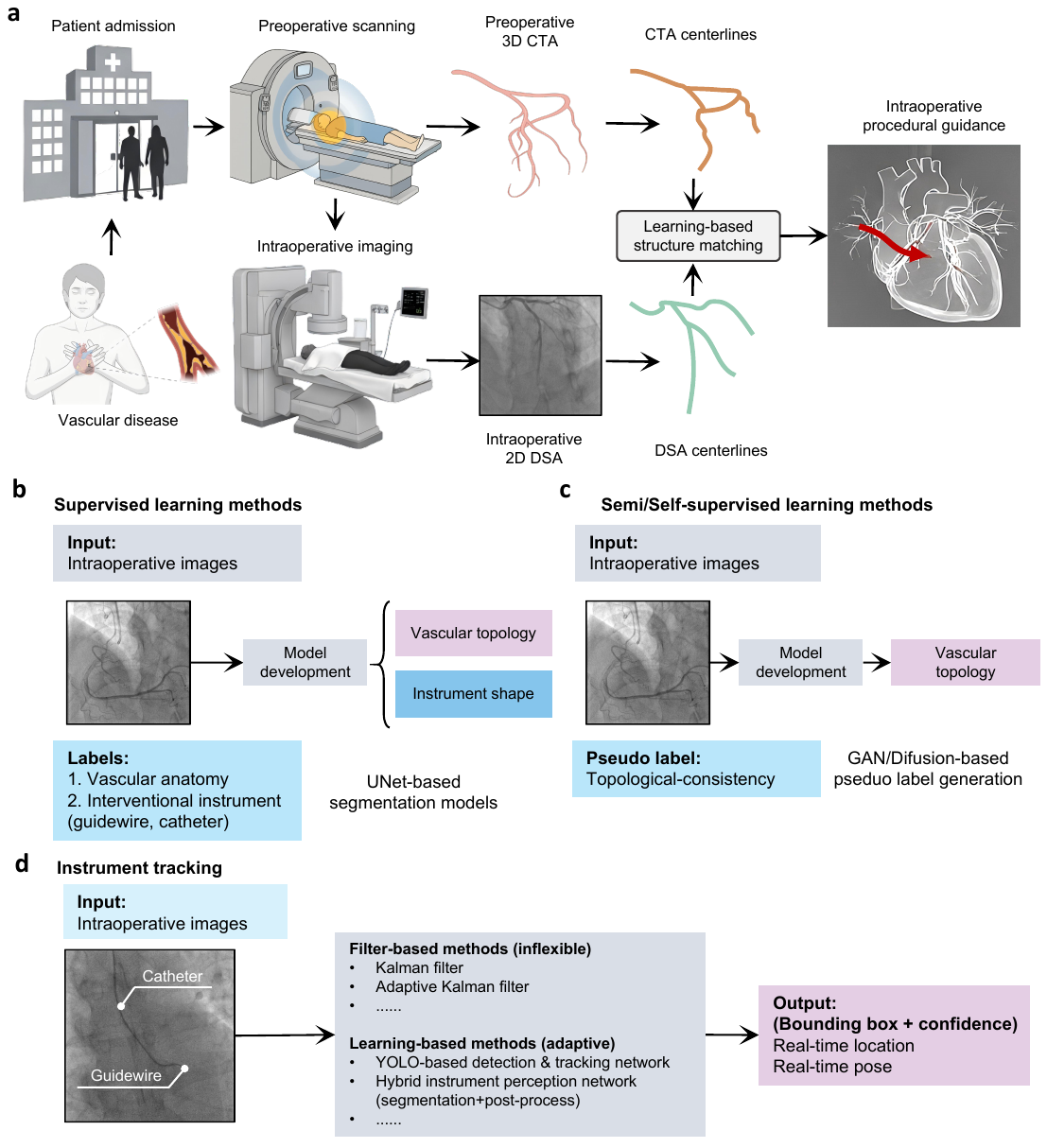}\vspace{-0.5cm}
\caption{Perception tasks in endovascular procedures. (a) Intraoperative vessel segmentation from DSA images to obtain detailed vascular anatomical structures \cite{iyer2021angionet}. (b) Device detection and localization utilizing intraoperative DSA images, providing essential visual feedback for physicians and robotic systems \cite{zhou2020lightweight}. (c) Multi-modal image registration of preoperative Computed tomography angiography (CTA) and intraoperative DSA for enhanced navigation cite{song2024iterative}. (d) Vessel segmentation from intraoperative DSA images using a GAN-based architecture with contrastive learning \cite{ma2021self}. (e) Endpoint-based CFKD-Net with multi-branch feature aggregation for weakly-supervised segmentation \cite{ma2023towards}. (f) Diffusion-based method for progressive vessel structure refinement \cite{kim2024c}. (g) Fast recurrent attention network for simultaneous mono- and dual-guidewire segmentation and tracking \cite{zhou2020lightweight}. (h) Coordinate regression-based deep learning for automated catheter detection \cite{aghasizade2023coordinate}.  Images reproduced from \cite{iyer2021angionet, zhou2020lightweight, song2024iterative, ma2021self, kim2024c, ma2023towards} and \cite{aghasizade2023coordinate}.\vspace{-0.55cm}} 
\label{fig:perception_tasks}
\end{figure*}

Digital Subtraction Angiography serves as the dominant intraoperative imaging modality due to its capacity for high-temporal-resolution vessel visualization, whereas preoperative Computed Tomography Angiography provides detailed three-dimensional anatomical context at the expense of radiation exposure and spatial registration complexity. Integration of these complementary modalities theoretically addresses inherent limitations such as depth ambiguity in projection imaging and temporal latency in volumetric acquisition. However, seamless fusion remains technically challenging due to respiratory motion, cardiac pulsatility, and contrast-dependent tissue appearance variations. Advances in deep learning have improved segmentation accuracy and device localization under controlled experimental conditions, yet systematic evaluations across diverse patient populations, imaging protocols, and institutional settings remain sparse. Critical gaps persist in understanding failure modes under anatomical edge cases, such as heavily calcified vessels, extreme tortuosity, or congenital anomalies, where algorithm performance degrades substantially but is infrequently reported.

This section organizes perception techniques according to their clinical function: vascular anatomy understanding for navigation and lesion targeting, interventional device perception for manipulation feedback, and multi-modal integration for enhanced spatial awareness. For each domain, the analysis emphasizes methodological trends, quantifies performance disparities across supervision paradigms and validation settings, and critically evaluates barriers to clinical deployment, including dataset bias, inconsistent benchmarking practices, and underreporting of computational requirements and failure cases. Table \ref{tab:clinical_vessel_seg_methods} consolidates quantitative performance metrics, enabling focus on interpretive synthesis rather than enumerative description.

\vspace{-0.4cm}

\subsection{Vascular Anatomy Understanding for Navigation and Lesion Localization}

Automated vascular segmentation supports robotic navigation by extracting vessel centerlines, bifurcation topologies, and stenotic regions that inform path planning algorithms and lesion targeting strategies. Accurate segmentation enables construction of patient-specific vascular roadmaps that guide autonomous device advancement while minimizing collision risk and procedural time. However, clinical deployment faces multiple obstacles: motion artifacts from cardiac contraction and respiratory excursion degrade image quality; projection ambiguity in two-dimensional fluoroscopy obscures vessel overlap; and annotation scarcity limits supervised learning approaches, particularly for rare anatomical variants or pathological conditions underrepresented in training datasets. Methodological evolution reflects tension between accuracy maximization through dense supervision and practical scalability through annotation-efficient paradigms.

Supervised learning methods, predominantly variants of the UNet architecture augmented with attention mechanisms, temporal aggregation, and multi-scale feature fusion, achieve the highest reported segmentation performance on benchmark datasets with comprehensive expert annotations. Representative examples include spatial-temporal frameworks incorporating motion compensation~\cite{wang2020coronary}, adversarial training for boundary refinement~\cite{shi2020uenet}, and recursive attention networks emphasizing vessel centerline extraction~\cite{han2022recursive, zhang2023centerline}. These approaches demonstrate Dice coefficients in the range of 0.83 to 0.90 on curated datasets with consistent imaging protocols and well-defined anatomical targets. However, performance deteriorates markedly in the presence of severe calcifications, which obscure vessel boundaries; tortuous anatomies, where projection overlap complicates segmentation; and low-contrast acquisitions, which reduce signal-to-noise ratios. Despite high accuracy under favorable conditions, these methods exhibit limited generalization to unseen patient populations, imaging equipment variations, or acquisition protocols differing from training distributions. Furthermore, clinical translation is hindered by the scarcity of large-scale, multi-institutional datasets with standardized annotations, leading to overfitting and algorithmic brittleness when deployed outside the originating institution.

Weakly-supervised and semi-supervised strategies emerged as responses to annotation bottlenecks, leveraging pseudo-labeling, consistency regularization, and adversarial self-supervision to reduce dependency on pixel-level expert annotations. Notable implementations include curriculum learning frameworks that progressively refine noisy labels~\cite{zhang2020weakly}, generative adversarial networks with contrastive objectives for feature discrimination~\cite{ma2021self}, and diffusion-based methods that model segmentation as iterative denoising~\cite{kim2024c}. These approaches typically achieve Dice coefficients approximately 0.05 to 0.15 lower than fully supervised counterparts but require substantially fewer annotated examples, thereby improving scalability across diverse datasets. Critical limitations include sensitivity to pseudo-label quality, which can propagate errors and destabilize training; difficulty capturing fine vascular structures such as distal branches below typical imaging resolution; and inconsistent performance across anatomical regions, with larger vessels segmented more reliably than peripheral vasculature. The clinical value proposition of reduced annotation burden is partially offset by increased algorithmic complexity and the need for careful hyperparameter tuning to balance pseudo-label confidence and model flexibility.

Clinically oriented developments prioritize robustness to real-world variability, real-time inference latency, and compatibility with diverse imaging hardware. Approaches incorporating geometric priors such as vessel width constraints~\cite{song2022coronary, yang2024segmentation}, edge-aware loss functions~\cite{guo2024edge}, and transfer learning from related anatomical domains~\cite{nobre2023coronary} attempt to encode domain knowledge that mitigates data scarcity. Among these, the multi-center validation by Menezes et al.~\cite{nobre2023coronary} represents a rare example of cross-institutional evaluation, demonstrating Dice coefficients exceeding 0.94 on consortium data but revealing performance degradation of approximately 10 to 15 percent in obese patients and cases with severe pathology. This variability underscores a fundamental challenge: benchmark performance on curated datasets often overstates clinical readiness, as real-world deployment encounters patient heterogeneity, imaging variability, and pathological edge cases inadequately represented in training cohorts. Moreover, few studies report inference latency, hardware requirements, or failure recovery strategies, limiting assessment of integration feasibility within time-critical clinical workflows. Regulatory pathways for autonomous perception algorithms remain underdeveloped, with most systems confined to research prototypes lacking the safety validation and quality management systems required for clinical certification.

\subsubsection{Available Public Datasets}

Reproducibility and benchmarking in vascular segmentation are fundamentally constrained by the scarcity of large-scale, publicly accessible datasets with expert annotations, diverse patient demographics, and standardized acquisition protocols. Privacy regulations and institutional data governance policies restrict sharing of clinical imaging, while annotation costs and inter-observer variability complicate ground-truth generation. Consequently, most studies rely on proprietary datasets that preclude external validation and hinder comparative evaluation. Existing public resources include the X-ray Angiography Coronary Artery Disease dataset~\cite{ma2021self}, comprising 1621 training frames and 126 test angiograms with weak annotations suitable for semi-supervised learning; Database X-ray Coronary Angiograms~\cite{liu2022full}, providing 134 and 30 expert-annotated sequences for multi-dataset validation; Automated Segmentation of Coronary Arteries~\cite{gharleghi2022automated}, offering 60 Computed Tomography Angiography cases with expert annotations for three-dimensional segmentation; Coronary Computed Tomography Angiography-200~\cite{yang2024segmentation}, containing 200 annotated volumes from radiological archives; and Coronary Artery Disease Detection by Invasive Coronary Angiography~\cite{wu2024denver}, aggregating over 6000 images for unsupervised approaches. Despite these contributions, dataset heterogeneity in imaging protocols, resolution, contrast administration, and pathology representation limits cross-study comparisons. Annotation inconsistencies, including differing definitions of vessel boundaries and treatment of overlapping structures, further complicate benchmark interpretation. Establishment of standardized evaluation protocols, including specification of preprocessing pipelines, performance metrics, and clinical relevance criteria, remains an urgent priority for the field.

\vspace{-0.3cm}

\subsection{Interventional Devices Perception in Endovascular Procedures}

Real-time tracking of guidewires and catheters, which function as continuum robots with nonlinear deformation characteristics~\cite{kwok2022soft}, is essential for closed-loop robotic control, collision avoidance, and operator situational awareness. Traditional sensor-embedded approaches, such as Fiber Bragg Grating arrays~\cite{shi2016shape}, introduce fabrication complexity, failure modes from sensor degradation, and constraints on device miniaturization. Vision-based perception offers a non-invasive alternative, extracting device pose and shape directly from intraoperative imaging without hardware modification. These methods support kinematic state estimation~\cite{almanzor2023static, ha2022shape}, force inference from deformation~\cite{feng2023study, brumfiel2024image}, and predictive modeling for motion planning~\cite{ma2025shape, huang2024predicting}. However, challenges arise from low signal-to-noise ratios in fluoroscopic images, occlusion by anatomical structures or contrast medium, and computational latency constraints imposed by real-time control requirements. Performance evaluations reveal a trade-off between model complexity and inference speed, with recurrent architectures achieving high accuracy at the expense of increased latency, while lightweight designs sacrifice precision for real-time capability.

\subsubsection{Guidewire Detection and Tracking}

Guidewire tracking underpins vascular navigation by providing continuous feedback on device configuration relative to target anatomy, enabling closed-loop control strategies that adjust insertion depth, rotation, and steering torque. Accurate tracking is particularly critical in tortuous vessels and bifurcation regions where guidewire buckling or vessel wall contact may precipitate complications. Deep learning approaches, predominantly based on recurrent convolutional architectures with attention mechanisms, address temporal consistency and long-range dependencies inherent in sequential fluoroscopic imaging. Representative methods include pyramid attention networks~\cite{zhou2020pyramid}, lightweight multi-scale frameworks~\cite{zhou2020lightweight}, and unified architectures for simultaneous multi-guidewire detection~\cite{li2021unified}. Reported performance metrics, such as Dice coefficients exceeding 0.94 for single-guidewire scenarios, suggest high accuracy under idealized conditions. However, these results are typically obtained on small, single-institution datasets with controlled imaging parameters and limited anatomical diversity. Multi-guidewire scenarios, relevant for complex interventions involving simultaneous access routes, exhibit performance degradation of approximately 5 to 10 percent due to occlusion, overlap, and ambiguity in endpoint association.

Probabilistic approaches incorporating uncertainty quantification~\cite{li2021gaussian} and kinematic-pose optimization~\cite{ravigopal2023real} attempt to address tracking failures by propagating measurement confidence through temporal filters and imposing mechanical consistency constraints. These methods improve robustness to intermittent occlusions and low-contrast frames but introduce computational overhead that may exceed real-time budgets on standard clinical workstations. Sim-to-real transfer strategies~\cite{wen2024generalizing}, leveraging synthetic data generation to augment limited clinical training sets, demonstrate modest improvements in cross-domain generalization but remain sensitive to discrepancies in simulated versus actual imaging characteristics, including noise models, scatter artifacts, and geometric calibration errors. Critical gaps include absence of failure-mode analysis, such as behavior under complete occlusion or rapid device motion; limited reporting of computational latency and hardware specifications; and lack of multi-center validation across diverse imaging equipment and operator techniques. These deficiencies obscure practical deployment feasibility and hinder comparative evaluation across studies.

\subsubsection{Catheter Detection and Tracking}

Catheter tracking enables precise control of therapeutic delivery, such as stent placement or embolic agent injection, by localizing device tips and shaft configurations relative to target lesions. Compared to guidewires, catheters exhibit larger diameters and more complex geometries, including radiopaque markers and variable stiffness profiles, which facilitate detection but complicate shape estimation. Methods range from lightweight segmentation networks~\cite{gherardini2020catheter, zhou2020frr} prioritizing inference speed to sophisticated coordinate regression frameworks~\cite{aghasizade2023coordinate} optimizing endpoint localization accuracy. Self-supervised approaches exploiting temporal optical flow consistency~\cite{ranne2024cathflow} reduce annotation requirements but introduce sensitivity to motion artifacts and contrast bolus dynamics, which violate brightness constancy assumptions underlying optical flow estimation. Integration with robotic kinematics through Bayesian filtering~\cite{ma2020dynamic} combines vision-based measurements with model predictions, achieving sub-millimeter localization errors in benchtop phantoms. However, translation to clinical settings encounters challenges from respiratory and cardiac motion, which induce non-rigid deformations not captured by kinematic models; overlapping anatomical structures that occlude catheter segments; and variability in fluoroscopic imaging angles that alter apparent device geometry.

Comparative evaluation across guidewire and catheter tracking reveals similar accuracy ranges but divergent failure modes. Guidewires, being thinner and more flexible, are prone to boundary detection errors and false positives from vessel edges or artifact streaks, whereas catheters face challenges in distinguishing shaft segments from background structures and accurately estimating curvature along tortuous paths. Despite claimed advantages of learning-based methods over classical image processing, few studies provide head-to-head comparisons under equivalent conditions, and performance improvements often reflect dataset-specific tuning rather than fundamental algorithmic superiority. Moreover, real-time requirements impose constraints on model capacity, favoring architectures with fewer parameters and shallower depth, which limits representational power and generalization capability. Clinical integration necessitates not only high detection accuracy but also robustness to diverse imaging protocols, graceful degradation under challenging conditions, and transparent failure signaling to operators, attributes infrequently evaluated in current literature.

\subsection{Multi-modal Registration and Integration}

Multi-modal image registration aligns complementary imaging data, such as preoperative three-dimensional Computed Tomography Angiography volumes with intraoperative two-dimensional Digital Subtraction Angiography sequences, to provide integrated spatial context for navigation and decision-making. Successful registration enables overlay of detailed anatomical roadmaps onto live fluoroscopy, facilitating target localization, trajectory planning, and assessment of device placement accuracy~\cite{song2024dynaweightpnp, song2024iterative, wu2022car, song2024vascularpilot3d}. However, registration is complicated by modality-specific artifacts, such as metal-induced streaking in Computed Tomography and motion blur in Digital Subtraction Angiography; physiological motion from respiration and cardiac pulsation, which induce non-rigid deformations between acquisitions; and contrast-dependent intensity variations that confound similarity metrics. Traditional intensity-based and feature-based registration methods struggle with these challenges, motivating adoption of learning-based approaches that implicitly model domain-specific transformations.

Deep learning registration frameworks~\cite{wu2022car, jaganathan2023self, huang2024real} leverage convolutional neural networks and spatial transformer modules to predict alignment parameters or deformation fields directly from image pairs. These methods achieve registration errors in the range of 1 to 3 millimeters on controlled datasets, representing substantial improvement over classical techniques. Notable implementations include deformable registration networks incorporating biomechanical priors~\cite{wu2022car}, self-supervised frameworks exploiting cycle-consistency constraints~\cite{jaganathan2023self}, and two-stage pipelines combining global rigid alignment with local deformable refinement~\cite{huang2024real}. Reinforcement learning formulations~\cite{liu2024udcr} model registration as sequential decision-making, optimizing transformation parameters through reward signals based on overlap metrics, demonstrating feasibility of policy-based search strategies for complex alignment tasks. Despite these advances, clinical translation faces obstacles from computational demands exceeding real-time budgets; sensitivity to initialization, which can trap optimization in local minima; and brittleness to anatomical changes between preoperative and intraoperative states, such as tissue deformation from device insertion or vasospasm. Furthermore, validation predominantly relies on retrospective data with simulated perturbations rather than prospective evaluation in live procedures, leaving uncertainty regarding registration accuracy under actual clinical conditions including patient motion, image quality degradation, and workflow time constraints.

Dynamic roadmapping~\cite{ma2020dynamic}, which maintains registration throughout procedures despite motion and device manipulation, represents a particularly challenging application requiring real-time tracking of deformations and occlusions. Approaches combining Bayesian filtering with learned image features attempt to balance accuracy and computational efficiency, but performance degrades in the presence of rapid motion, large deformations, or loss of landmark visibility. Critical evaluation reveals that reported error metrics, typically based on landmark distances or overlap coefficients, may not correlate with clinical utility, as registration errors below visual detection thresholds can still impact procedural outcomes when aggregated over multiple decision points. Moreover, absence of failure-mode characterization, such as behavior under complete landmark occlusion or registration divergence, limits understanding of reliability boundaries. Standardized benchmarks incorporating diverse anatomies, realistic motion profiles, and clinically relevant error tolerances are needed to facilitate method comparison and establish performance thresholds for clinical deployment.

\begin{figure*}[!htbp]
    \centering
    \includegraphics[width=0.90\textwidth, keepaspectratio]{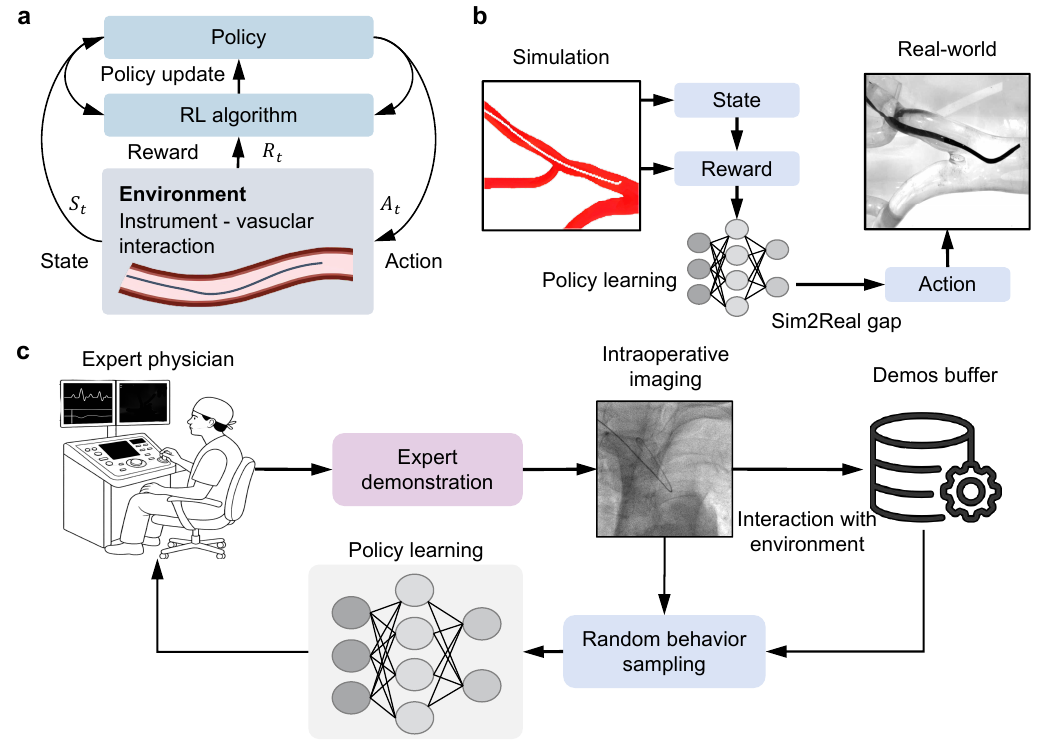}
    \vspace{-0.5cm}
    \caption{Learning frameworks for robotic-assisted endovascular procedures can be categorized as follows: (a) The traditional reinforcement learning (RL) \cite{singh2022reinforcement} framework for endovascular navigation involves a robotic agent learning through direct interaction with the vascular environment. This approach utilizes target networks and experience buffers to optimize catheter/guidewire manipulation policies \cite{robertshaw2023artificial, karstensen2022learning, song2022learning, tian2023ddpg, scarponi2024zero}. (b) The demonstration-based learning framework incorporates expert cardiologist demonstrations to guide the learning of robotic manipulation policies. This method enables more efficient acquisition of complex endovascular navigation skills \cite{li2023casog, zhou2022learning, li2024model}. (c) The human-in-the-loop learning framework integrates expert cardiologist supervision into the robotic training process. This approach combines automated learning with clinical expertise to enhance safety and efficacy in endovascular procedures \cite{jianu2024cathsim, jianu2024autonomous}.\vspace{-0.5cm}}
    \label{fig:RL}
\end{figure*}

\begin{table*}[!hbbp]
\centering
\caption{Summary of Data-Driven Control Methods for Endovascular Intervention Robots: Clinical Applications and Challenges}
\renewcommand{\arraystretch}{1.3}
\begin{tabular}{p{3.0cm}p{2.5cm}p{1.0cm}p{6.0cm}p{3.0cm}}
\toprule
\textbf{Clinical Task} & \textbf{Authors (Year)} & \textbf{Setting} & \textbf{Method and Key Features} & \textbf{Performance Metrics} \\
\midrule
\multirow{6}{*}{\parbox{3.0cm}{Guidewire and Catheter Navigation in Static Environments \\  \textit{Challenge:} Precise control in complex vascular structures with minimal error}} & Kweon \textit{et al.} (2021) \cite{kweon2021deep} & Physical & DRL-based navigation (PPO, DDPG, SAC); \textit{Key Features:} CNN-based state encoding, custom reward design, discrete/continuous action spaces, and safety-aware policies for stable navigation & Task completion rate, path accuracy, control stability, safety metrics \\
\cmidrule{2-5} & Yang \textit{et al.} (2022) \cite{yang2022guidewire} & Simulation & DRL-based navigation (PPO, DDPG, SAC); \textit{Key Features:} CNN-based state encoding, custom reward design, discrete/continuous action spaces, and safety-aware policies for stable navigation & Task completion rate, path accuracy, control stability, safety metrics \\
\cmidrule{2-5} & Li \textit{et al.} (2024) \cite{li2024model} & Simulation & DRL-based navigation (PPO, DDPG, SAC); \textit{Key Features:} CNN-based state encoding, custom reward design, discrete/continuous action spaces, and safety-aware policies for stable navigation & Task completion rate, path accuracy, control stability, safety metrics \\
\cmidrule{2-5} & Wang \textit{et al.} (2024) \cite{wang2024autonomous} & Simulation & DRL-based navigation (PPO, DDPG, SAC); \textit{Key Features:} CNN-based state encoding, custom reward design, discrete/continuous action spaces, and safety-aware policies for stable navigation & Task completion rate, path accuracy, control stability, safety metrics \\
\cmidrule{2-5} & Karstensen \textit{et al.} (2022) \cite{karstensen2022learning} & Both & Vision-based autonomous navigation; \textit{Key Features:} Path planning and real-time control for venous system navigation & Success rate, navigation precision, safety metrics \\
\cmidrule{2-5} & Jolaei \textit{et al.} (2021) \cite{jolaei2021toward} & Physical & Learning-based kinematic control; \textit{Key Features:} Neural network controller and tendon-driven model for cardiac ablation & Control accuracy, position error, response time \\
\midrule
\multirow{4}{*}{\parbox{3.0cm}{Navigation in Dynamic Environments \\ \textit{Challenge:} Aadaptation to physiological variations and motion}} & Scarponi \textit{et al.} (2024) \cite{scarponi2024autonomous,scarponi2024zero} & Simulation & Autonomous navigation with zero-shot and sim-to-real strategies; \textit{Key Features:} Domain adaptation, policy transfer, and robust control for dynamic environments & Navigation accuracy, adaptability, robustness \\
\cmidrule{2-5} & Yao \textit{et al.} (2025) \cite{yao2025sim2real} & Both & Autonomous navigation with zero-shot and sim-to-real strategies; \textit{Key Features:} Domain adaptation, policy transfer, and robust control for dynamic environments & Navigation accuracy, adaptability, robustness \\
\cmidrule{2-5} & Liu \textit{et al.} (2024) \cite{liu2024image} & Simulation & Vision-based RL navigation; \textit{Key Features:} Multi-modal fusion and image processing for real-time control & Navigation accuracy, precision, real-time performance \\
\cmidrule{2-5} & Robertshaw \textit{et al.} (2024) \cite{robertshaw2024autonomous} & Simulation & DRL-based end-to-end navigation; \textit{Key Features:} Vision-based control and safety constraints for intra-vascular navigation & Task success rate, navigation efficiency, safety metrics \\
\midrule
\multirow{3}{*}{\parbox{3.0cm}{Skill Learning and Transfer for Interventional Procedures \\ \textit{Challenge:} Reproducing expert skills with minimal training data}} & Zhou \textit{et al.} (2022) \cite{zhou2022learning} & Simulation & Skill learning with conservative actor-critic and behavior cloning; \textit{Key Features:} Feature extraction, smooth gradient policies, and experience replay for skill transfer & Learning efficiency, task accuracy, convergence speed \\
\cmidrule{2-5} & Li \textit{et al.} (2023) \cite{li2023casog} & Both & Skill learning with conservative actor-critic and behavior cloning; \textit{Key Features:} Feature extraction, smooth gradient policies, and experience replay for skill transfer & Learning efficiency, task accuracy, convergence speed \\
\cmidrule{2-5} & Chi \textit{et al.} (2020) \cite{chi2020collaborative} & Physical & GAN-based imitation learning; \textit{Key Features:} WGAN architecture and collaborative control for trajectory generation & Success rate, learning efficiency, real-time performance \\
\cmidrule{2-5} & Mei \textit{et al.} (2024) \cite{mei2024transferring} & Simulation & Hierarchical actor-critic learning; \textit{Key Features:} Knowledge transfer and hierarchical learning for autonomous delivery & Success rate, learning speed, decision accuracy \\
\midrule
\multirow{3}{*}{\parbox{3.0cm}{Simulation-based Training and Control \\ \textit{Challenge:} Sim2real gap for safe training}} & Jianu \textit{et al.} (2024) \cite{jianu2024cathsim} & Simulation & Open-source simulator; \textit{Key Features:} Physics-based modeling and real-time simulation for training environments & Simulation fidelity, real-time execution, training effectiveness \\
\cmidrule{2-5} & Tian \textit{et al.} (2023) \cite{tian2023ddpg} & Simulation & DDPG-based virtual training; \textit{Key Features:} Virtual environment and transfer learning for catheter navigation & Navigation accuracy, real-time performance, generalization \\
\cmidrule{2-5} & Dreyfus \textit{et al.} (2022) \cite{dreyfus2022simulation} & Simulation & RL-based magnetic navigation; \textit{Key Features:} Magnetic field control, path optimization, and safety constraints & Position error, navigation time, safety index \\
\bottomrule
\end{tabular}
\label{tab:control_methods_clinical}
\vspace{1em}
\footnotesize \textit{Note:} Performance metrics are qualitative as reported in the original papers, focusing on task success, accuracy, efficiency, and safety. Setting classifications: Simulation (virtual environments only), Physical (real phantoms or ex-vivo/in-vivo), Both (combination or sim-to-real transfer).
\end{table*}

\section{Learning-Based Control in Robotic-Assisted Endovascular Procedures} \label{Control}

Endovascular instruments, functioning as continuum robots with infinite degrees of freedom and nonlinear mechanical coupling between actuation and tip motion~\cite{dupont2022continuum}, present fundamental control challenges that resist analytical modeling. Classical control strategies predicated on rigid-body kinematics fail in tortuous vascular pathways where frictional contact, elastic energy storage, and fluid-structure interaction dominate device behavior~\cite{berrueta2024materializing}. Anatomical variability across patient populations~\cite{pore2023autonomous} and limited intraoperative sensing~\cite{nazeer2024rl, chen2024data} further complicate model-based approaches, motivating adoption of data-driven control paradigms~\cite{ni2023data}. Learning-based methods, including reinforcement learning, imitation learning, and hybrid frameworks, offer the potential to capture complex input-output relationships through experience~\cite{laschi2023learning, kwok2022soft}, adapt policies to individual anatomies, and refine control strategies iteratively. Proponents argue these approaches enhance precision, reduce operator workload, and improve procedural safety and efficiency~\cite{zhang2022robotic}. However, clinical deployment remains scarce, raising questions regarding the gap between reported algorithmic performance in controlled settings and practical robustness in real-world procedures.

Methodological trends reflect migration from hardware-intensive real-world validation toward simulation-dominated development pipelines with subsequent sim-to-real transfer. This transition is driven by ethical constraints on in vivo experimentation, high costs of clinical data acquisition, and practical advantages of simulation for systematic algorithm evaluation under controlled perturbations. While simulation accelerates development and enables high-throughput exploration of algorithmic variants, it introduces fidelity gaps that may not fully capture contact mechanics, tissue deformation, imaging artifacts, and physiological variability encountered clinically. Consequently, performance claims derived from simulation require critical interpretation, particularly when transfer validation is absent or limited to benchtop phantoms with idealized geometries. Furthermore, inconsistent reporting of experimental conditions, including details of reward function design, hyperparameter selection, computational infrastructure, and failure case analysis, hinders reproducibility and comparative evaluation across studies. Table \ref{tab:control_methods_clinical} consolidates reported methods, validation settings, and performance metrics, revealing heterogeneity in evaluation protocols that complicates direct comparison and assessment of translational readiness.

\begin{figure*}[!htbp]
\centering
\includegraphics[width=0.96\textwidth]{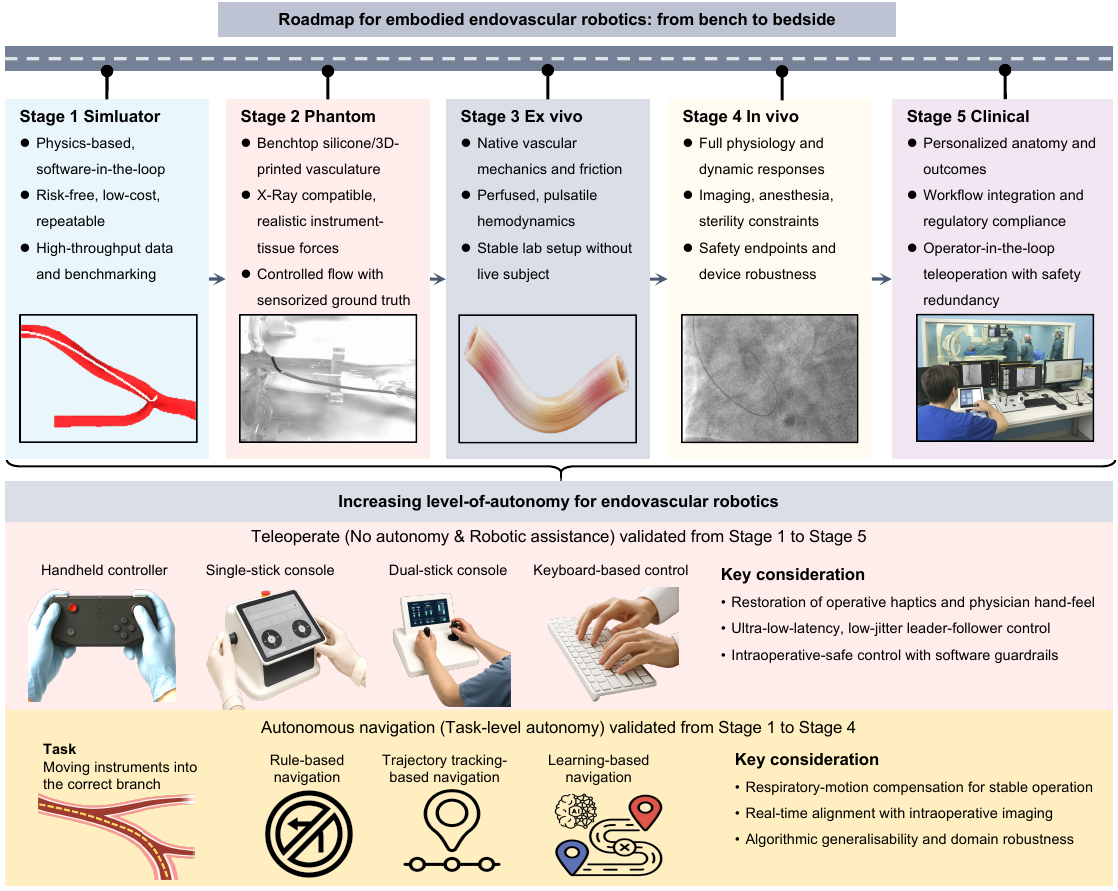}\vspace{-0.3cm}
\caption{Roadmap for achieving trustworthy Embodied Intelligence in endovascular robotics, illustrating the coupled progression of autonomous capabilities and safety validation. Here, EI is tangibly defined as the robot's ability to form a closed perception-action loop, using real-time sensory data to execute adaptive physical maneuvers. As the complexity of these embodied behaviors (the level of intelligence) increases, the demands for safety and reliability become exponentially more stringent. \textbf{A Staged Validation Framework.} This framework is designed to rigorously verify these escalating reliability requirements. The pathway progresses from low-risk, high-throughput algorithm development in simulators (Stage 1), to controlled physical testing of device-tissue interaction in phantoms (Stage 2). It then moves to more realistic ex vivo (Stage 3) and in vivo (Stage 4) environments that capture full physiological dynamics and require evaluation against critical safety endpoints (e.g., vessel perforation, dissection rates). The final stage, clinical deployment (Stage 5), demands the highest level of proven reliability, integrating workflow compatibility and regulatory compliance under strict operator-in-the-loop supervision. \textbf{A Safety-Centric Autonomy Ladder.} Progression up this ladder is contingent upon meeting the reliability standards validated in the corresponding stages above. As more of the procedure is performed by the embodied agent, the burden of proof for safety shifts to the system's demonstrated intelligence and robustness. \textbf{Teleoperation (Level 0, no autonomy)} requires baseline safety features like software guardrails and force limits. \textbf{Assistive Guidance (Level 1)} introduces AI-powered perception, demanding reliability in tasks like vessel segmentation and instrument tracking. \textbf{Shared Control (Level 2)} requires the system to prove its reliability in executing simple, predictable sub-tasks (e.g., straight-line advancement). \textbf{Supervised Autonomy (Level 3)}, such as task-level navigation, demands a much higher level of intelligence and reliability, including proven algorithmic generalisability and robustness under out-of-distribution conditions (e.g., unexpected anatomical variants), which must be systematically validated through the preceding stages.}
\label{fig:Roadmap}
\vspace{-0.55cm}
\end{figure*}

Real-world validation, encompassing benchtop phantoms, ex vivo tissue models, in vivo animal studies, and clinical trials, provides the most direct assessment of clinical applicability but encounters substantial barriers. Ethical review processes, regulatory compliance, high costs of clinical access, and safety concerns constrain experimentation, particularly for novel algorithms lacking established safety records. Consequently, real-world evaluations are typically confined to small sample sizes, homogeneous patient or animal populations, and controlled procedural scenarios that may not represent the full spectrum of clinical variability. Early efforts, including deep reinforcement learning for coronary navigation~\cite{kweon2021deep} and behavior cloning for skill acquisition~\cite{zhou2022learning}, demonstrate the feasibility of learning-based control in physical settings and report improvements in path accuracy compared to baseline heuristics. However, these studies acknowledge limitations in generalization across diverse anatomies, robustness to unexpected events such as device buckling or vessel spasm, and scalability given data scarcity. Furthermore, th absence of multi-center validation, long-term follow-up, and head-to-head comparison with expert operators limits the interpretation of clinical utility. Positioning within the validation roadmap (Figure \ref{fig:Roadmap}) situates these efforts at Stage 4 (in vivo) with preliminary steps toward Stage 5 (clinical), where safety endpoints, workflow integration, and regulatory approval dominate translation milestones.

Simulation-based development leverages physics engines, finite element models, and procedural generation to create virtual environments for algorithm training and evaluation. Platforms such as CathSim~\cite{jianu2024cathsim} provide open-source frameworks incorporating vascular geometry reconstruction, device mechanics, and simplified hemodynamics, enabling reproducible benchmarking and ablation studies. Reinforcement learning implementations in simulation~\cite{li2023casog, yang2022guidewire} demonstrate policy convergence and task completion under idealized conditions, with training throughput orders of magnitude higher than real-world data collection. Proponents argue that simulation accelerates iterative development, facilitates systematic sensitivity analysis, and enables exploration of rare failure modes difficult to capture experimentally. However, simulation fidelity gaps introduce domain shift: contact friction models may not capture tissue adhesion and lubrication effects; vascular geometry simplifications omit anatomical fine structure and pathological variations; and imaging simulation often neglects scatter, beam hardening, and patient-specific noise characteristics. Consequently, policies optimized in simulation may perform suboptimally or fail when deployed in real environments, a phenomenon termed the reality gap. Methodological advances in domain randomization, visual domain adaptation, and physics-informed simulation attempt to mitigate this gap, but transfer validation remains inconsistent across studies. In the roadmap framework, simulation corresponds to Stage 1 (simulators), providing low-risk, high-throughput experimentation essential for early-stage development but insufficient alone for clinical translation.

Sim-to-real transfer strategies attempt to bridge fidelity gaps through domain adaptation, policy distillation, and hybrid modeling combining learned and analytical components. Techniques include domain randomization of simulation parameters to encourage policy robustness~\cite{tian2023ddpg}, visual domain translation between synthetic and real imaging modalities~\cite{scarponi2024zero, scarponi2024autonomous}, and fine-tuning on limited real-world data following pre-training in simulation~\cite{yao2025sim2real}. Reported improvements in transfer performance are often qualitative or evaluated on narrow task distributions, with limited analysis of failure modes arising from unmodeled dynamics or distribution shifts. For instance, zero-shot transfer methods~\cite{scarponi2024zero} demonstrate task completion in phantom environments but exhibit degraded success rates when anatomical complexity increases or imaging conditions deviate from training assumptions. Hybrid approaches integrating multi-modal perception~\cite{wang2024autonomous, liu2024image} and constrained reinforcement learning for safety-critical navigation~\cite{dreyfus2022simulation} achieve modest gains in robustness and generalization, yet quantitative comparisons across transfer strategies are scarce. Critical evaluation reveals that claims of performance improvements, such as increased success rates or reduced navigation errors, often reflect algorithmic tuning specific to experimental setups rather than fundamental advances generalizable across platforms and procedures. Moreover, underreporting of negative results, failed transfer attempts, and algorithmic limitations biases the literature toward optimistic assessments, obscuring true state-of-the-art performance boundaries.

Imitation learning and demonstration-based approaches seek to leverage expert knowledge by training policies to replicate human operator behavior, thereby incorporating domain expertise and reducing exploration requirements compared to trial-and-error reinforcement learning. Methods include behavior cloning from recorded demonstrations~\cite{zhou2022learning}, inverse reinforcement learning to infer reward functions from expert trajectories~\cite{li2023casog}, and generative adversarial imitation frameworks~\cite{chi2020collaborative}. These approaches demonstrate faster convergence and improved sample efficiency in controlled settings but face challenges from distribution mismatch between expert demonstrations and autonomous policy execution, covariate shift leading to compounding errors, and difficulty capturing implicit operator strategies not observable in kinematic trajectories. Furthermore, expert variability and inconsistency in demonstration quality introduce noise that can degrade learned policies. Transfer learning frameworks~\cite{mei2024transferring} attempt to decompose skills into reusable components, but generalization across procedural contexts remains limited. Critical assessment reveals that imitation learning performance depends heavily on demonstration dataset quality, coverage of relevant state-action distributions, and algorithmic capacity to extrapolate beyond observed examples, factors inadequately characterized in current literature.

\section{Challenges} \label{Challenges}
Data-driven approaches in endovascular robotic systems offer considerable promise, yet several critical challenges persist that need to be addressed to fully harness their potential. These challenges lie within the technical, ethical, and systemic domains, areas in which substantial progress is necessary to advance key capabilities and clinical adoption.

\vspace{-0.4cm}
\subsection{Scarcity of High-Quality Imaging Data}
Despite the extreme importance of having high-quality reference data in endovascular imaging, current literature reveals surprisingly little research on this. The inherent complexity of vascular structures, combined with significant variability in patient presentation and imaging conditions, necessitates robust gold standards and extensive, well-curated datasets for reliable AI development. However, the acquisition of such data is problematic. Endovascular imaging data often requires specialized expertise for annotation, and the dynamic nature of arteries, influenced by cardiac and respiratory motion, adds complexity. Empirical evidence suggests that in vascular imaging, data quality is often more crucial than algorithmic sophistication in determining model performance \cite{ma2024segment, huang2024segment, jin2025aortic}. This dependency suggests that resource allocation should prioritize data-curation infrastructure over purely algorithmic innovation. The challenge transcends data volume, extending to representativeness; current datasets often exhibit geographic, demographic, and institutional biases that limit model generalizability. Furthermore, the temporal dimension of interventions necessitates datasets that capture dynamic procedural sequences, instrument–tissue interactions, and physiological responses, not merely static anatomy. A lack of standardized quality metrics for these datasets fundamentally hinders systematic evaluation and comparability across studies. Coordinated efforts to establish standardized protocols for data collection, annotation, and sharing are therefore essential to capture diverse pathologies and anatomical variations.

\vspace{-0.4cm}
\subsection{Absence of a Standardized, Large-Scale Benchmark}
Recent advancements in large-scale benchmarking, such as the Behavior-1K dataset \cite{li2024behavior}, have significantly advanced performance evaluation in general robotics. In contrast, endovascular robotics is constrained by a lack of domain-specific benchmarks that encapsulate the intricacies of human vasculature. A central finding of this review is that this absence of benchmarks has led to a pronounced heterogeneity across studies, which manifests as inconsistent experimental settings, metrics, and validation protocols. This fragmentation not only makes meaningful synthesis difficult but also fundamentally slows scientific progress and clinical translation. This lack of standardization stems not only from data collection difficulties but also from an absence of consensus on clinically meaningful metrics, the challenge of standardizing anatomical variability, and ethical constraints on data sharing. Unlike discrete robotic tasks, success in endovascular interventions is multifaceted, encompassing procedural efficiency, radiation safety, and long-term clinical outcomes that manifest post-intervention. Although emerging efforts like CathAction \cite{huang2024cathaction} are valuable, current benchmarks predominantly focus on isolated subtasks, failing to capture the synergistic complexity of complete procedures or their inherent stochasticity \cite{zhang2024generalist}.

Addressing this gap requires a benchmarking framework tailored to endovascular robotics. Such a framework must balance standardization with flexibility to accommodate institutional variations. It should incorporate metrics for uncertainty quantification, out-of-distribution detection, and worst-case performance to characterize safety margins, alongside a hierarchical evaluation structure ranging from phantom studies to high-fidelity physiological models. This infrastructure would facilitate consistent, quantitative assessments and foster a cycle of iterative improvement.

Insights from broader robotics highlight both opportunities and challenges. While large-scale simulation platforms (for example, NVIDIA OMNIVERSE) and surgical benchmarking frameworks (for example, SurRoL \cite{xu2021surrol}) demonstrate the power of scalable, simulation-based development, their direct transfer is complicated. These domains often benefit from more structured task definitions, richer sensory feedback, and greater hardware standardization than endovascular procedures. Adapting methodologies from rigid-link manipulators to continuum devices requires addressing underactuation and complex environmental constraints unique to intravascular navigation. Moreover, the safety-critical nature of endovascular interventions mandates validation standards that exceed those in many other robotic fields, demanding rigorous protocols to account for catastrophic failure modes like vessel perforation.

\vspace{-0.4cm}
\subsection{Underdevelopment of a Foundational Model for Endovascular Procedures}
Despite the success of transformer-based architectures in medicine and robotics \cite{mazurowski2023segment, ma2023towardsf}, their integration into endovascular procedures is nascent. The specific characteristics of endovascular imaging pose distinct perception challenges for foundation models. The low signal-to-noise ratio of fluoroscopy, combined with motion artefacts from cardiac and respiratory cycles, creates a difficult environment compared with the natural image domains where these models have excelled. Additionally, the requirement for real-time inference imposes computational constraints that are incompatible with many large-scale transformer architectures. The integration of such models with multi-modal inputs, including preoperative imaging and intra-operative fluoroscopy, also remains underdeveloped.

The aforementioned general scarcity of high-quality data becomes a particularly acute barrier when considering the development of foundation models, which are predicated on pre-training at an unprecedented scale. While transformers excel when trained on vast corpora, current endovascular datasets are often several orders of magnitude smaller than those that have driven success in other domains. Moreover, the challenge is compounded by a profound domain shift that is not merely visual but also semantic; endovascular procedures involve continuous control objectives, which contrasts sharply with the discrete classification or segmentation tasks on which many foundation models are pre-trained. The proprietary nature of clinical data and privacy regulations, discussed previously, further constrain the large-scale collaborative data aggregation efforts necessary for foundation model development. Therefore, leveraging these powerful architectures will require a concerted, community-wide effort to establish the specific type of large-scale, multi-modal data infrastructure that is currently absent.

\vspace{-0.4cm}
\subsection{Lack of Workflow Analysis for Endovascular Procedures}
The absence of a unified approach to workflow recognition in endovascular procedures impedes the development of higher-level autonomy. In contrast to laparoscopic surgery, where benchmarks like Autolaparo \cite{wang2022autolaparo} exist, comparable resources for endovascular interventions are lacking. This disparity arises partly from inherent procedural differences; laparoscopic surgery often follows more stereotyped workflows with clearer visual landmarks, whereas endovascular interventions exhibit greater variability driven by patient-specific anatomy and angiographic findings. The predominantly two-dimensional fluoroscopic view further complicates recognition compared with the three-dimensional vision in laparoscopy.

Accurate workflow recognition is paramount for managing instrument sequences and anticipating procedural transitions. Beyond instrument management, workflow recognition could enable context-aware system behaviours, such as automated imaging adjustments or predictive alerts for complications. However, developing such systems requires addressing fundamental questions regarding the optimal granularity of phase segmentation, the management of concurrent tasks, and generalization across procedural variations. A central unresolved tension exists between creating standardized models for reliable recognition and maintaining the flexibility to accommodate operator and institutional preferences.

\vspace{-0.4cm}
\subsection{Scalability of Perception and Learning-Based Control}
Scalability is a pivotal challenge for perception and control systems in the complex, dynamic vascular environment. A unique perception challenge arises from the coupling between device manipulation and tissue deformation, as the system simultaneously alters the environment it is observing. For learning-based control, the distribution shift between training data and clinical deployment is a critical vulnerability. This shift manifests not only in anatomy but also in equipment and imaging protocols. A crucial safety gap is the frequent absence of robust uncertainty quantification, leading to systems that may fail silently and exhibit overconfidence in novel scenarios. This lack of generalizability significantly reduces effectiveness in real-world environments.

To address these issues, advanced perception algorithms that integrate multi-modal imaging and deep learning are needed. Promising control solutions include meta-learning frameworks for rapid patient-specific adaptation and hybrid architectures that combine learned policies with model-based safety guarantees. However, the safety of online adaptation must be carefully validated. Consequently, validation at each stage of development (per Fig.~\ref{fig:Roadmap}) should incorporate explicit out-of-distribution testing to systematically assess generalization boundaries and failure modes.

\vspace{-0.4cm}
\subsection{The Translation Gap: From Technical Validation to Clinical Deployment}
A substantial gap persists between technical validation and clinical adoption, with few systems progressing beyond early-stage phantom or simulation studies. This gap is not solely regulatory but systemic. First, academic incentives often favour novel proof-of-concept demonstrations over the iterative engineering required for clinical-grade reliability. Second, the considerable capital investment needed for commercialization and regulatory approval is deterred by an uncertain return on investment in a specialized market. Third, a critical barrier is the frequent disconnect between technical development and clinical need, where systems are designed without continuous end-user engagement, leading to poor workflow integration. Bridging this gap necessitates systemic change, including funding mechanisms that support long-term translational research, collaborative frameworks that unite engineers, clinicians, and industry partners from project inception, and a broader recognition of the timeline and complexity inherent in translating medical robotic systems to clinical practice.

\vspace{-0.4cm}
\subsection{Ethical and Legal Concerns}
The integration of autonomous systems into medicine poses profound ethical and legal challenges. Current ethical discourse in endovascular robotics often overlooks the fundamental tensions between autonomous capabilities and the core values of medical practice. Determining liability is complicated by the distributed responsibility among manufacturers, developers, and clinicians, a problem exacerbated by the opacity of many machine learning models. The demand for explainable AI (XAI) reflects a clinical necessity for understanding system reasoning, yet a technical trade-off exists between model interpretability and performance. Trustworthiness requires not just post-hoc explanations but also comprehensive characterization of system behaviour and competence boundaries, a feature lacking in many current XAI methods.

The complexity of AI also challenges traditional informed consent; explaining algorithmic uncertainty, the implications of model updates, and the use of collective data is difficult. Similarly, while federated learning offers a potential solution to data privacy, it introduces new security vulnerabilities and does not fully resolve privacy concerns. The high cost and specialized nature of these systems also raise equity concerns, with the potential to create a two-tiered system of care and for algorithms to underperform in underrepresented demographic groups. Fundamentally, the goal should not be full autonomy but a carefully considered division of labour, pursuing graduated autonomy that delegates well-defined tasks while preserving human oversight for strategic and ethical decisions.

\vspace{-0.3cm}
\subsection{Integration with Existing Clinical Workflows}
The integration of robotic systems into clinical workflows is a critical barrier to adoption. Clinician reluctance is often rooted in the "black-box" nature of autonomous systems. This reluctance stems from concerns about accountability, potential skill degradation, and the loss of the tactile feedback and intuitive control that define manual expertise. The introduction of robotic systems also fundamentally alters team dynamics, necessitating new communication protocols and role definitions. This aligns with Stage 5 in our proposed roadmap, which emphasizes interoperability and operator-in-the-loop supervision.

Interoperability with existing medical infrastructure is another major hurdle. This challenge is compounded by a medical device ecosystem characterized by proprietary interfaces and a lack of open standards, creating commercial and regulatory barriers beyond technical incompatibility. Effective integration must address not only data exchange but also control authority and safety interlocks between disparate systems. True integration must span the entire procedural timeline, from pre-procedural planning to post-procedural documentation, without creating administrative burdens. The optimal level of assistance in this augmentation paradigm requires careful calibration to avoid inducing operator complacency or skill atrophy, suggesting a need for systems that can adapt their level of autonomy.

Finally, training and education are essential. Training paradigms must be adapted from manual techniques, with a greater emphasis on simulation, failure mode management, and protocols for transitioning to manual control. The development of standardized curricula and certification is essential. Furthermore, institutional learning curves, which encompass team coordination and workflow optimization, represent a significant and often underestimated investment required for successful adoption.

\section{Future Directions}
\subsection{Panvascular Embodied Intelligence for Generalization of Endovascular Intervention}
Panvascular medicine offers a unified strategy for treating systemic vascular disease \cite{khera2024transforming} . Correspondingly, panvascular EI describes a new class of physically grounded AI agents capable of adaptive, expert-level decision-making across diverse vascular territories. To establish clear definitional boundaries, EI in this context is a paradigm where the robotic system, predicated on sensory information such as real-time fluoroscopy, forms a closed logical loop to guide instrument manipulation. This approach is distinct from both purely rule-based systems, which lack adaptability, and purely teleoperated systems, where all intelligence resides with the human operator. Modern EI incorporates learning-based methods, enabling the agent to fuse experiential knowledge with autonomous learning capabilities to form sophisticated operational skills. This "prior experience plus real-time closed-loop control" model allows the agent to execute precise physical actions based on both prior learning and current sensory data. Moreover, the introduction of high-fidelity simulation enables the agent to accumulate this experience through high-throughput trial-and-error learning in a safe, controlled environment.

The ultimate objective of Panvascular EI is not to replace the clinician, but to upskill them and democratize access to high-quality care. A generalizable EI agent can function as a highly effective tool that shortens the learning curve for complex procedures and elevates the proficiency of operators in low-volume centers. This capability directly addresses a major challenge in modern healthcare: enabling community hospitals to safely offer cutting-edge treatments that are currently concentrated in major academic medical centers.

Among all surgical modalities, endovascular intervention presents uniquely favourable conditions for realizing such high-level autonomy. The constrained luminal anatomy simplifies the perception and control problem relative to open soft-tissue surgery. Additionally, the standardization of interventional tools and the availability of continuous imaging provide a rich foundation for a data-driven, embodied approach. These characteristics reduce environmental uncertainty and simplify the perception-to-action mapping, which are core prerequisites for autonomous operation.

Building on this foundation, panvascular EI agents require several key capabilities. These include multimodal perception for comprehensive situation awareness, compositional reasoning to decompose procedures into reusable skill primitives, and continual learning for patient-specific adaptation. Crucially, they also require robust uncertainty-aware decision-making. As interventions will inevitably present unanticipated anatomical variations or events, the embodied agent must be able to recognize when it is operating outside its domain of competence. This necessitates formal safety assurance methods, such as runtime monitoring for out-of-distribution detection and the integration of conservative, physics-based fallback mechanisms to ensure patient safety. The autonomy ladder in Fig.~\ref{fig:Roadmap}~provides a curriculum for the safe, high-reliability deployment of these agents, progressing from assistive guidance to supervised task-level autonomy, with human oversight remaining paramount.

\vspace{-0.4cm}

\subsection{Developing Foundation Models and Benchmarks Tailored for Endovascular Procedures}
Foundation models promise to enable robust perception and adaptive control in endovascular procedures, provided they are trained on large, heterogeneous, multimodal corpora. Such datasets must encompass not only imaging (for example, DSA, CTA) but also device kinematics, haemodynamics, and procedural reports to capture the full context of an intervention. Model architectures must then address domain-specific requirements, including temporal modelling for long-range dependencies, multi-scale spatial reasoning, and real-time inference latencies, all while incorporating robust uncertainty quantification for safety.

The impact of these models, however, hinges on standardized, clinically grounded benchmarks. Such a framework is the primary solution to the critical challenge of methodological heterogeneity identified in this review, providing the unified evaluation criteria needed to move the field beyond its current fragmented state. Following precedent from challenges such as EndoVis and DRIVE, benchmark design should mirror the clinical workflow. For vascular segmentation, metrics must include not only the centerline F1 score but also topological connectivity, thin-branch sensitivity, and stenosis quantification error. For device perception, evaluation should report tip error, temporal stability, and latency under stress tests like occlusion. For learning-based control, benchmarks must standardize tasks such as branch cannulation and lesion crossing, reporting not only success rates but also safety-proximal measures like vessel wall contact events and recovery from defined perturbations. Reliability must be rigorously assessed through institution-held-out splits, explicit out-of-distribution tracks, and calibrated uncertainty metrics. A staged Vascular Autonomy Benchmark, directly mapping to the validation framework in Fig.~\ref{fig:Roadmap}, is recommended. This requires consortium-backed data aggregation and open evaluation toolkits to accelerate the development of trustworthy, generalizable models ready for prospective validation.

\subsection{Patient-Specific Adaptation and Embodied Workflow Integration}
A critical future direction is the adaptation of embodied agents to individual patients. While the concept of fine-tuning models on patient-specific preoperative data is appealing, its implementation for an embodied agent presents profound challenges that transcend standard machine learning. For an EI system, personalization is not merely an update of an algorithm; it is the modification of a physical policy that will directly interact with that patient's unique anatomy. The notion of "online adaptation," where an embodied agent learns during a live procedure, introduces the risk of catastrophic failure. An erroneous update based on an intraoperative anomaly could lead to unsafe physical actions, such as vessel perforation. This highlights a safety-critical version of the stability-plasticity dilemma: the embodied agent must be plastic enough to adapt to the patient's anatomy, yet stable enough to guarantee its physical actions remain within safe boundaries. A more pragmatic pathway may involve offline embodied personalization, where a patient-specific "digital twin" is used to simulate and validate the adapted policy before it is ever deployed in the physical robot.

Furthermore, workflow integration for an embodied agent is a problem of human-robot teaming, not just software interoperability. The clinician must not only trust the AI's "mind" but also its "body"—the physical robot. Integrating a physically acting, semi-autonomous agent requires new protocols for shared control, intention communication, and risk management. If the agent's actions are not predictable or its intentions are unclear, it can increase the clinician's cognitive load and undermine trust, leading to manual override and defeating the purpose of autonomy. True workflow integration, therefore, demands a seamless cognitive and physical partnership, where the clinician can intuitively supervise and direct the embodied agent as a reliable member of the surgical team.

\subsection{Regulatory and Ethical Frameworks for Embodied AI}
The physical embodiment of AI in endovascular robotics fundamentally reshapes the regulatory and ethical landscape. Current regulatory models, designed for static medical devices or purely informational AI (like diagnostic software), are inadequate for embodied agents whose actions have direct physical consequences. The core challenge is that an embodied AI system is a dynamic, integrated entity of software and hardware. An update to its learning-based policy could instantly alter its physical behavior, potentially turning a previously safe robot into an unsafe one. This creates a critical regulatory gap, demanding novel frameworks for the continuous verification and validation of embodied systems, not just their initial software components.

The "black-box" problem is also amplified by embodiment. The challenge transcends understanding a recommendation; it requires understanding the genesis of a physical action. In the event of an adverse outcome, attributing liability is extraordinarily complex. Was the cause a flaw in the AI's policy, a mechanical failure in the robot, a sensor misreading, or an appropriate but misapplied action by the supervising clinician? This distributed responsibility across the entire embodied system (software, hardware, and human) makes forensic analysis and legal accountability exceptionally difficult. Future ethical and legal frameworks must therefore address the system as a whole, perhaps requiring auditable logs that record not just the AI's decision, but also the sensory data that prompted it and the resulting physical execution.

This leads to the fundamental ethical question of appropriate autonomy for an embodied agent. Rather than pursuing full autonomy, a more ethically defensible goal is a collaborative paradigm where the physician's role evolves \cite{dupont2025grand} . The expert clinician transitions from a master of manual dexterity to a high-level supervisor of an intelligent, embodied tool. In this model, the human retains ultimate moral and clinical responsibility, making strategic decisions while leveraging the robot's superhuman precision or tireless execution. This human-on-the-loop approach respects the unique capabilities of both the human and the embodied agent, providing a pragmatic and ethically sound path for the integration of intelligent robotics into high-stakes medical procedures.

\section{Conclusion}
The integration of data-driven methodologies into endovascular robotics represents a pivotal moment for minimally invasive procedures. This systematic review has analyzed the convergence of advanced robotics and AI through the unifying lens of EI. Our analysis reveals a field rich with algorithmic innovation in perception and control, yet simultaneously constrained by systemic challenges. We identify the profound heterogeneity in validation standards as a core finding, which fragments progress and impedes meaningful comparison. Moreover, a persistent translation gap highlights a fundamental disconnect between technical development and clinical reality, rooted not only in practical barriers but also in a paradigm mismatch between mimicking human skill and leveraging machine-native capabilities.

Based on this analysis, this review argues for a strategic reorientation of the field's ultimate objective. Rather than the pursuit of complete, unsupervised autonomy, the more pragmatic and ethically sound goal is a sophisticated human-robot collaboration paradigm. This vision focuses on developing intelligent, embodied agents that serve to augment, not replace, the clinician. It foresees an evolution of the expert interventionalist's role from a master of manual dexterity to a high-level supervisor who interprets complex data and makes strategic decisions, while the robotic system executes tasks with superhuman precision or consistency. It is this model of augmented intelligence that offers the most promising path to safer, more effective, and more accessible endovascular care.

Realizing this vision requires a concerted effort focused on several critical research priorities. The establishment of comprehensive, standardized benchmarks is paramount to overcoming the field's current fragmentation. The development of robust, generalizable foundation models, trained on large-scale, multi-modal datasets, is necessary to power next-generation perception and control. Finally, continued research into explainable AI (XAI) and formal safety verification is essential for building clinical trust and navigating the complex regulatory landscape. The achievement of these objectives necessitates sustained, interdisciplinary collaboration between clinicians, engineers, and regulatory bodies, providing a principled foundation to guide the next generation of intelligent endovascular robotics.

\vspace{-0.25cm}
\printbibliography
\balance

\end{document}